\documentclass[11pt,letterpaper]{article}
\usepackage[letterpaper]{geometry}
\usepackage{acl2012}
\usepackage{times}
\usepackage{latexsym}
\usepackage{amsmath}
\usepackage{amssymb}
\usepackage{graphicx}
\usepackage{multirow}
\usepackage{subfig}
\usepackage{url}
\usepackage{paralist}
\usepackage{latexsym}
\usepackage{amsmath}
\usepackage{color}
\usepackage{graphicx}
\usepackage{booktabs}
\usepackage{subfig}
\usepackage{url}
\usepackage{tabularx}
\usepackage{verbatim}
\usepackage{CJK}
\usepackage{multicol}
\usepackage{textcomp}
\usepackage{balance}

\makeatletter
\newcommand{\@BIBLABEL}{\@emptybiblabel}
\newcommand{\@emptybiblabel}[1]{}

\newcommand \TODO[1]{{\bf \color{red} \\TODO:#1.\\}}
\newcommand{\tabincell}[2]{\begin{tabular}{@{}#1@{}}#2\end{tabular}}
\makeatother
\usepackage[hidelinks]{hyperref}

\DeclareMathOperator*{\argmin}{arg\,min}

\usepackage{expl3}
\ExplSyntaxOn
\newcommand\latinabbrev[1]{
	\peek_meaning:NTF . {
		#1\@}%
	{ \peek_catcode:NTF a {
			#1.\@ }%
		{#1.\@}}}
\ExplSyntaxOff

\title{Modeling Past and Future for Neural Machine Translation}
\author{
	Zaixiang Zheng\footnotemark[1] \\ Nanjing University \\ {\normalsize \em zhengzx@nlp.nju.edu.cn} \And
	Hao Zhou\footnotemark[1] \\ Toutiao AI Lab \\ {\normalsize \em zhouhao.nlp@bytedance.com} \And
	Shujian Huang \\ Nanjing University \\ {\normalsize \em huangsj@nlp.nju.edu.cn}
	\AND
	Lili Mou \\ University of Waterloo \\ {\normalsize \em doublepower.mou@gmail.com} \And
	Xinyu Dai  \\  Nanjing University \\ {\normalsize \em dxy@nlp.nju.edu.cn} \And
	Jiajun Chen \\ Nanjing University \\ {\normalsize \em chenjj@nlp.nju.edu.cn} \And
	Zhaopeng Tu \\ Tencent AI Lab \\ {\normalsize \em zptu@tencent.com}
}

\begin{document}
	\maketitle
	\renewcommand{\thefootnote}{\fnsymbol{footnote}}

	\begin{abstract}
		Existing neural machine translation systems do not explicitly model what has been translated and what has not during the decoding phase.
		To address this problem, we propose a novel mechanism that separates the source information into two parts: translated \textsc{Past} contents and untranslated \textsc{Future} contents, which are modeled by two additional recurrent layers. The \textsc{Past} and \textsc{Future} contents are fed to both the attention model and the decoder states, which provides NMT systems with the knowledge of translated and untranslated contents. Experimental results show that the proposed approach significantly improves the performance in Chinese-English, German-English, and English-German translation tasks. Specifically, the proposed model outperforms the conventional coverage model in terms of both the translation quality and the alignment error rate.\footnotemark[2]
	\end{abstract}

	\footnotetext[1]{Equal contributions.}
	\footnotetext[2]{Our code can be downloaded from \url{https://github.com/zhengzx-nlp/past-and-future-nmt}.}
	\renewcommand{\thefootnote}{\arabic{footnote}}
	\setcounter{footnote}{0}
	
	\section{Introduction}
	\label{sec:intro}
	
	Neural machine translation~(NMT) generally adopts an \textit{encoder-decoder} framework~\cite{D13-1176,D14-1179,sutskever2014sequence}, where the \textit{encoder} summarizes the source sentence into a \textit{source context vector}, and the \textit{decoder} generates the target sentence word-by-word based on the given source.
	During translation, the decoder implicitly serves several functionalities at the same time:
	\begin{compactenum}
		\item [1.] Building a language model over the target sentence for translation fluency (\textsc{Lm}).
		\item [2.] Acquiring the most relevant source-side information to generate the current target word (\textsc{Present}).
		\item [3.] Maintaining what parts in the source have been translated~(\textsc{Past}) and what parts have not~(\textsc{Future}).
	\end{compactenum}

	However, it may be difficult for a single recurrent neural network~(RNN) decoder to accomplish these functionalities simultaneously.
	A recent successful extension of NMT models is the \textit{attention} mechanism~\cite{bahdanau2014neural,D15-1166},
	which makes a soft selection over source words and yields an \textit{attentive vector} to represent the most relevant source parts for the current decoding state. In this sense, the attention mechanism separates the \textsc{Present} functionality from the decoder RNN, achieving significant performance improvement.
	
	In addition to \textsc{Present}, we address the importance of modeling \textsc{Past} and \textsc{Future} contents in machine translation.
	The \textsc{Past} contents indicate translated information, whereas the \textsc{Future} contents indicate untranslated information, both being crucial to NMT models, especially to avoid under-translation and over-translation~\cite{tu-EtAl:2016:P16-1}.
	Ideally, \textsc{Past} grows and \textsc{Future} declines during the translation process.
	However, it may be difficult for a single RNN to explicitly model the above processes.
	
	In this paper, we propose a novel neural machine translation system that explicitly models \textsc{Past} and \textsc{Future} contents with two additional RNN layers.
	The RNN modeling the \textsc{Past} contents (called \textsc{Past} layer) starts from scratch and accumulates the information that is being translated at each decoding step (i.e., the \textsc{Present} information yielded by attention). The RNN modeling the \textsc{Future} contents (called \textsc{Future} layer) begins with holistic source summarization, and subtracts the \textsc{Present} information at each step. The two processes are guided by proposed auxiliary objectives.
	Intuitively, the RNN state of the \textsc{Past} layer corresponds to source contents that have been translated at a particular step, and the RNN state of the \textsc{Future} layer corresponds to source contents of untranslated words. 
	At each decoding step, \textsc{Past} and \textsc{Future} together provide a full summarization of the source information.
	We then feed the \textsc{Past} and \textsc{Future} information to both the attention model and decoder states.
	In this way, our proposed mechanism not only provides coverage information for the attention model, but also gives a holistic view of the source information at each time.

	We conducted experiments on Chinese-English, German-English, and English-German benchmarks. Experiments show that the proposed mechanism yields 2.7, 1.7, and 1.1 improvements of BLEU scores in three tasks, respectively.
	In addition, it obtains an alignment error rate of 35.90\%, significantly lower than the baseline (39.73\%) and the coverage model (38.73\%) by \newcite{tu-EtAl:2016:P16-1}.
	We observe that in traditional attention-based NMT, most errors occur due to over- and under-translation, which is probably because the decoder RNN fails to keep track of what has been translated and what has not.
	Our model can alleviate such problems by explicitly modeling \textsc{Past} and \textsc{Future} contents.

	\section{Motivation}
	\label{sess:nmt}
	
	\begin{figure}[t]
		\centering
		\includegraphics[width=0.4\textwidth]{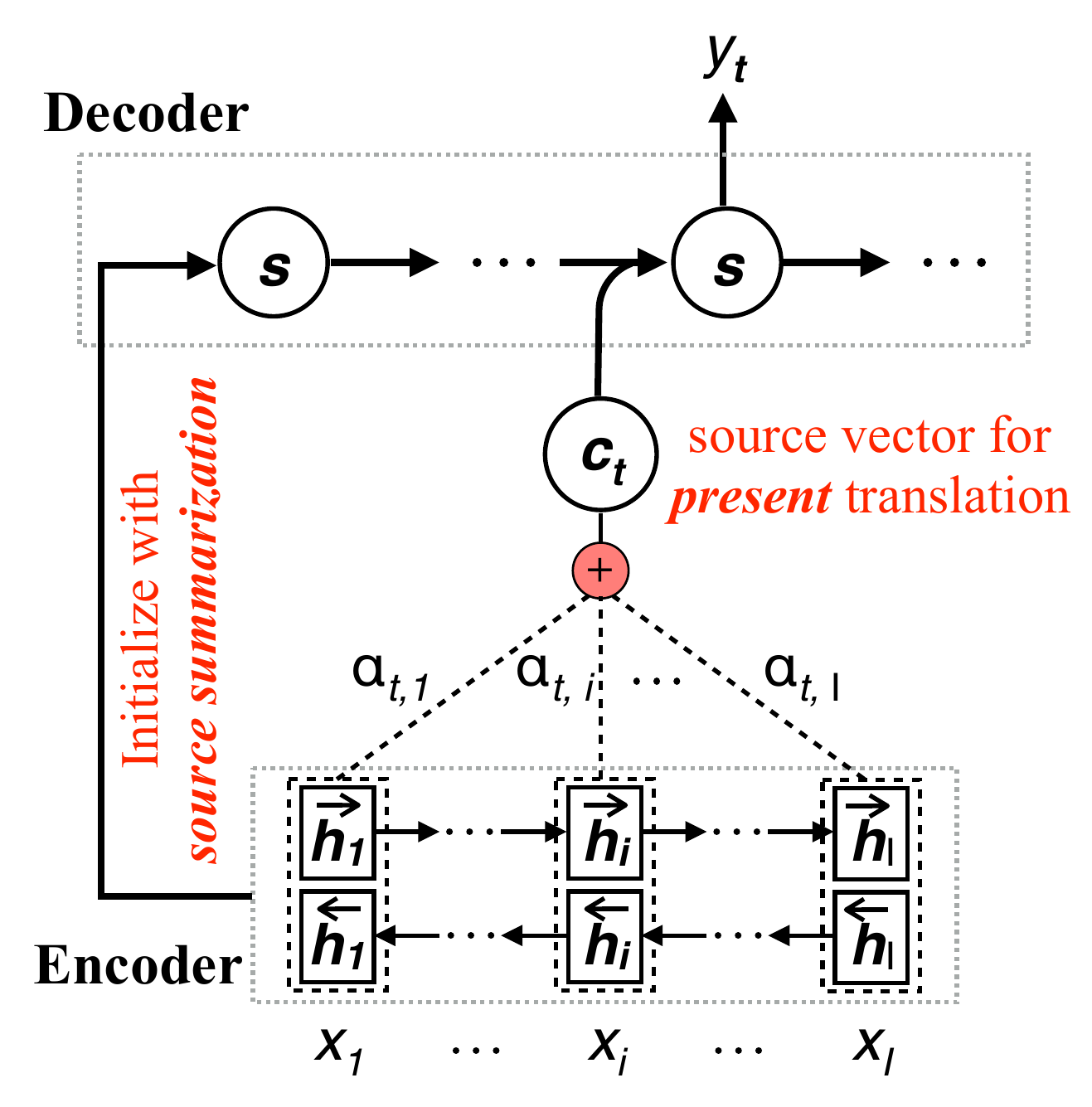}
		\caption{Architecture of attention-based NMT.}
		\label{fig:nmt}
	\end{figure}
	
	In this section, we first introduce the standard attention-based NMT, and then motivate our model by several empirical findings.

	The attention mechanism, proposed in \newcite{bahdanau2014neural}, yields a dynamic source context vector for the translation at a particular decoding step, modeling \textsc{Present} information as described in Section~\ref{sec:intro}. This process is illustrated in Figure~\ref{fig:nmt}.
	
	Formally, let ${\mathbf x}=\{x_1, \dots, x_{I}\}$ be a given input sentence. The encoder RNN---generally implemented as a bi-directional RNN~\cite{Schuster:1997:TSP}---transforms the sentence to a sequence of annotations with $\mathbf{h}_i=\big[\overrightarrow{\bf h}_i;\overleftarrow{\bf h}_i\big]$ being the annotation of $x_i$. ($\overrightarrow{\bf h}_i$ and $\overleftarrow{\bf h}_i$ refer to RNN's hidden states in both directions.)
	
	Based on the source annotations, another decoder RNN generates the translation by predicting a target word $y_t$ at each time step $t$:
	\begin{equation}\label{key}
	P(y_t|y_{<t},\mathrm{\mathbf{x}}) =\mathrm{softmax}(g(y_{t-1}, \mathbf{s}_t, \mathbf{c}_t))
	\end{equation}
	where $g(\cdot)$ is a non-linear activation, and $\mathbf{s}_t$ is the decoding state for time step $t$, computed by
	\begin{equation}\label{dec_state}
	\mathbf{s}_t=f(y_{t-1}, \mathbf{s}_{t-1}, \mathbf{c}_t)
	\end{equation}
	Here $f(\cdot)$ is an RNN activation function, e.g., the Gated Recurrent Unit \cite[GRU]{D14-1179} and Long Short-Term Memory~\cite[LSTM]{Hochreiter1997}.
	$\mathbf{c}_t$ is a vector summarizing relevant source information. It is computed as a weighted sum of the source annotations
	\begin{equation}
	\mathbf{c}_t = \sum_{i=1}^{I}{\alpha_{t,i} \cdot \mathbf{h}_i}
	\label{eqn-context}
	\end{equation}
	where the weights ($\alpha_{t,i}$ for $i=1\cdots,I$) are given by the attention mechanism:
	\begin{equation}
	\alpha_{t,i} = \mathrm{softmax}\big(a(\mathbf{s}_{t-1}, \mathbf{h}_i)\big)
	\label{eqn-alignment-probability}
	\end{equation}
	Here, $a(\cdot)$ is a scoring function measuring the degree to which the decoding state and source information match to each other.
	
	Intuitively, the attention-based decoder selects source annotations that are most relevant to the decoder state, based on which the current target word is predicted. In other words, $\mathbf{c}_t$ is some source information for the \textsc{Present} translation.
	
	\begin{CJK}{UTF8}{gbsn}
		\begin{table}[t]
			\centering
			\subfloat[Translation example. We highlight {\bf under-translated} words in bold and italicize \textit{over-translated} words.]{\begin{tabular}{m{25pt}|m{160pt}}
					\hline
					\textsc{Src} & 与此同时,他呼吁提高民事服务效率, 这也是鼓舞民心之举。\\
					\hline
					\textsc{Ref} &  in the meanwhile he calls for better efficiency in civil service , \textbf{which helps to promote people 's trust .}\\
					\hline
					\textsc{NMT} &  at the same time , he called for a higher \textit{efficiency} in civil service \textit{efficiency} .\\
					\hline
			\end{tabular}}\\
			\subfloat[Source summarization is not fully exploited by NMT decoder.]{
				\begin{tabular}{c|c}
					\hline
					Initialize Decoder States with \dots & BLEU\\
					\hline
					Source Summarization    & 35.13\\
					All-Zero Vector         & 35.01\\
					\hline
			\end{tabular}}
			\caption{Evidence shows that attention-based NMT fails to make full use of source information, thus losing the holistic picture of source contents.}
			\label{tab:bleu_init}
		\end{table}
	\end{CJK}
	
	The decoder RNN is initialized with the summarization of the entire source sentence $\big[\overrightarrow{\bf h}_I;\overleftarrow{\bf h}_1\big]$, given by
	\begin{equation}
	\mathbf{s}_0 = \tanh(W_s \big[\overrightarrow{\bf h}_I;\overleftarrow{\bf h}_1\big])
	\label{eqn-initialization}
	\end{equation}
	
	After we analyze existing attention-based NMT in detail, our intuition arises as follows.
	Ideally, with the source summarization in mind, after generating each target word $y_t$ from the source contents $\mathbf{c}_t$, the decoder should keep track of (1) translated source contents by accumulating $\mathbf{c}_t$, and (2) untranslated source contents by subtracting $\mathbf{c}_t$ from the source summarization.
	However, such information is not well learned in practice, as there lacks explicit mechanisms to maintain translated and untranslated contents.
	Evidence show that attention-based NMT still suffers from serious over- and under-translation problems~\cite{tu-EtAl:2016:P16-1,tu_etal:17}. Examples of under-translation are shown in Table~\ref{tab:bleu_init}a.
	
	Another piece of evidence also shows the decoder may lack a holistic view of the source information, explained as below. We conduct a pilot experiment by removing the initialization of the RNN decoder.
	If the ``holistic'' context is well exploited by the decoder, translation performance would significantly decrease without the initialization. As shown in Table~\ref{tab:bleu_init}b, however, translation performance only decreases slightly after we remove the initialization. This indicates NMT decoders do not make full use of source summarization ; that the
	initialization only helps the prediction at the beginning of the sentence.
	We attribute the vanishing of such signals to the overloaded use of decoder states (e.g., \textsc{Lm}, \textsc{Past}, and \textsc{Future} functionalities), and hence we propose to explicitly model the holistic source summarization by \textsc{Past} and \textsc{Future} contents at each decoding step.


	\begin{figure*}[t]
		\centering
		\includegraphics[width=0.75\textwidth]{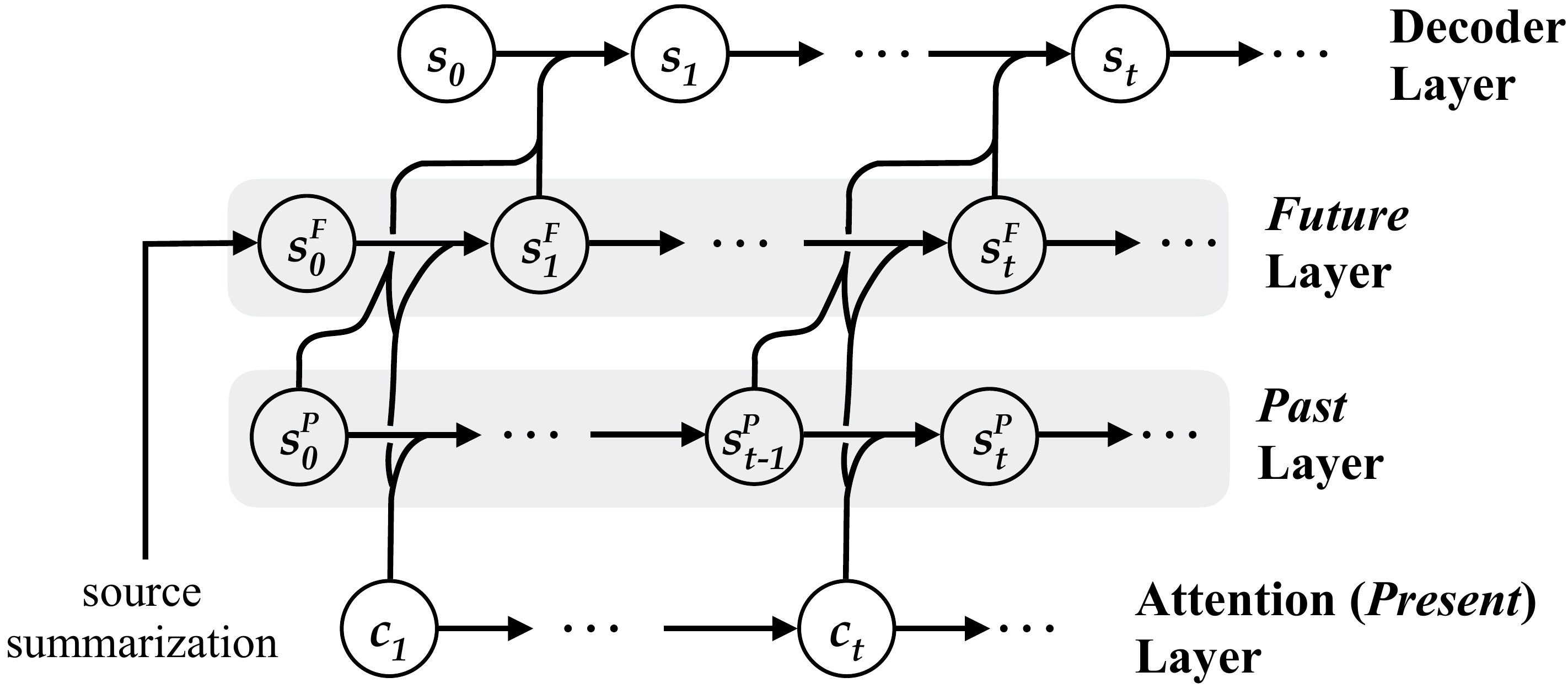}
		\caption{NMT decoder augmented with \textsc{Past} and \textsc{Future} layers.}
		\label{fig:tri-sRNN}
	\end{figure*}

	\section{Related Work}
	
	Our research is built upon an attention-based sequence-to-sequence model~\cite{bahdanau2014neural}, but is also related to coverage modeling, future modeling, and functionality separation. We discuss these topics in the following.
	
	\paragraph{Coverage Modeling.}
	\newcite{tu-EtAl:2016:P16-1} and \newcite{Mi2016} maintain a coverage vector to indicate which source words have been translated and which source words have not. These vectors are updated by accumulating attention probabilities at each decoding step, which provides an opportunity for the attention model to distinguish translated source words from untranslated ones.
	Viewing coverage vectors as a (soft) indicator of translated source contents, we take one step further following this idea. We model translated and untranslated source contents by directly manipulating the attention vector (i.e., the source contents that are being translated) instead of attention probability (i.e., the probability of a source word being translated).
	
	In addition, we explicitly model both translated (with \textsc{Past}-RNN) and untranslated (with \textsc{Future}-RNN) instead of using a single coverage vector to indicate translated source words. Another difference with \newcite{tu-EtAl:2016:P16-1} is that the \textsc{Past} and \textsc{Future} contents in our model are fed to not only the attention mechanism but also the decoder's states.

	In the context of semantic-level coverage,~\newcite{Wang:2016:EMNLP} propose a memory-enhanced decoder and ~\newcite{Meng:2016:COLING} propose a memory-enhanced attention model. Both implement the memory with a Neural Turing Machine~\cite{Graves:2014:arXiv}, in which the reading and writing operations are expected to erase translated contents and highlight untranslated contents. However, their models lack an explicit objective to guide such intuition, which is one of the key ingredients for the success in this work. In addition, we use two separate layers to explicitly model translated and untranslated contents, which is another distinguishing feature of the proposed approach.

	\paragraph{Future Modeling.}
	
	Standard neural sequence decoders generate target sentences from left to right, thus failing to estimate some desired properties in the future (e.g., the length of target sentence). To address this problem, actor-critic algorithms are employed to predict future properties~\cite{Li:2017:arXiv,Bahdanau:2017:ICLR}; in their models, an interpolation of the actor (the standard generation policy) and the critic (a value function that estimates the future values) is used for decision making.
	Concerning the future generation at each decoding step, \newcite{Weng:2017:EMNLP} guide the decoder's hidden states to not only generate the current target word, but also predict the target words that remain untranslated. Along the direction of future modeling, we introduce a \textsc{Future} layer to maintain the untranslated source contents, which is updated at each decoding step by subtracting the source content being translated (i.e., attention vector) from the last state (i.e., the untranslated source content so far).

	\paragraph{Functionality Separation.}
	Recent work has revealed that the overloaded use of representations makes model training difficult, and such problem can be alleviated by explicitly separating these functions~\cite{Reed2015Neural,Ba:2016:NIPS,Miller:2016:EMNLP,Gulcehre:2016:arXiv,Rocktaschel:2017:ICLR}. For example,~\newcite{Miller:2016:EMNLP} separate the functionality of look-up keys and memory contents in memory networks~\cite{Sukhbaatar:2015:NIPS}.
	~\newcite{Rocktaschel:2017:ICLR} propose a {\it key-value-predict} attention model, which outputs three vectors at each step: the first is used to predict the next-word distribution; the second serves as the key for decoding; and the third is used for the attention mechanism.
	In this work, we further separate \textsc{Past} and \textsc{Future} functionalities from the decoder's hidden representations.

	\begin{figure*}[t]
		\centering
		\subfloat[GRU]{
			\includegraphics[width=0.32\textwidth]{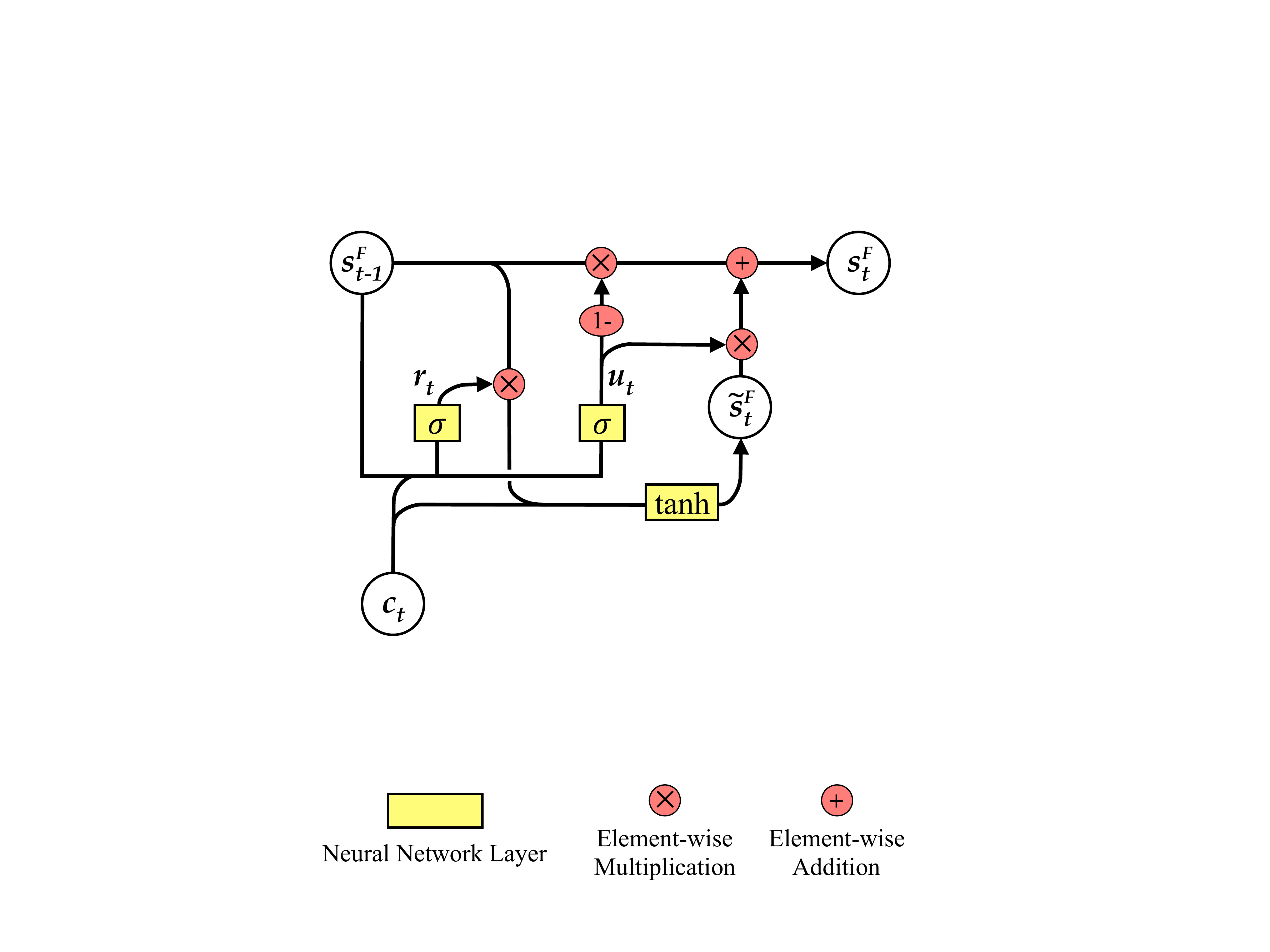}}
		\hfill
		\subfloat[GRU-$o$]{
			\includegraphics[width=0.32\textwidth]{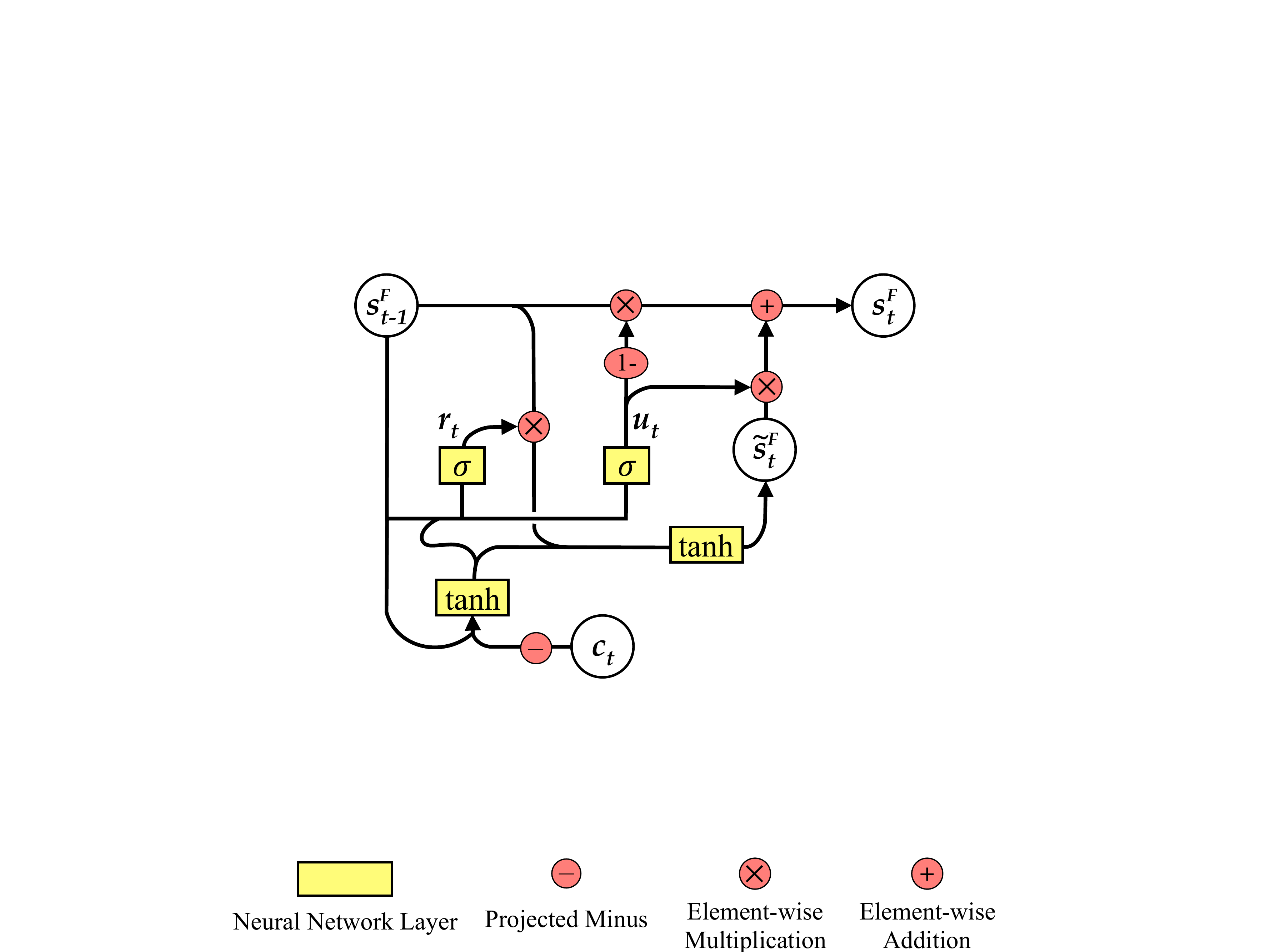}}
		\hfill
		\subfloat[GRU-$i$]{
			\includegraphics[width=0.32\textwidth]{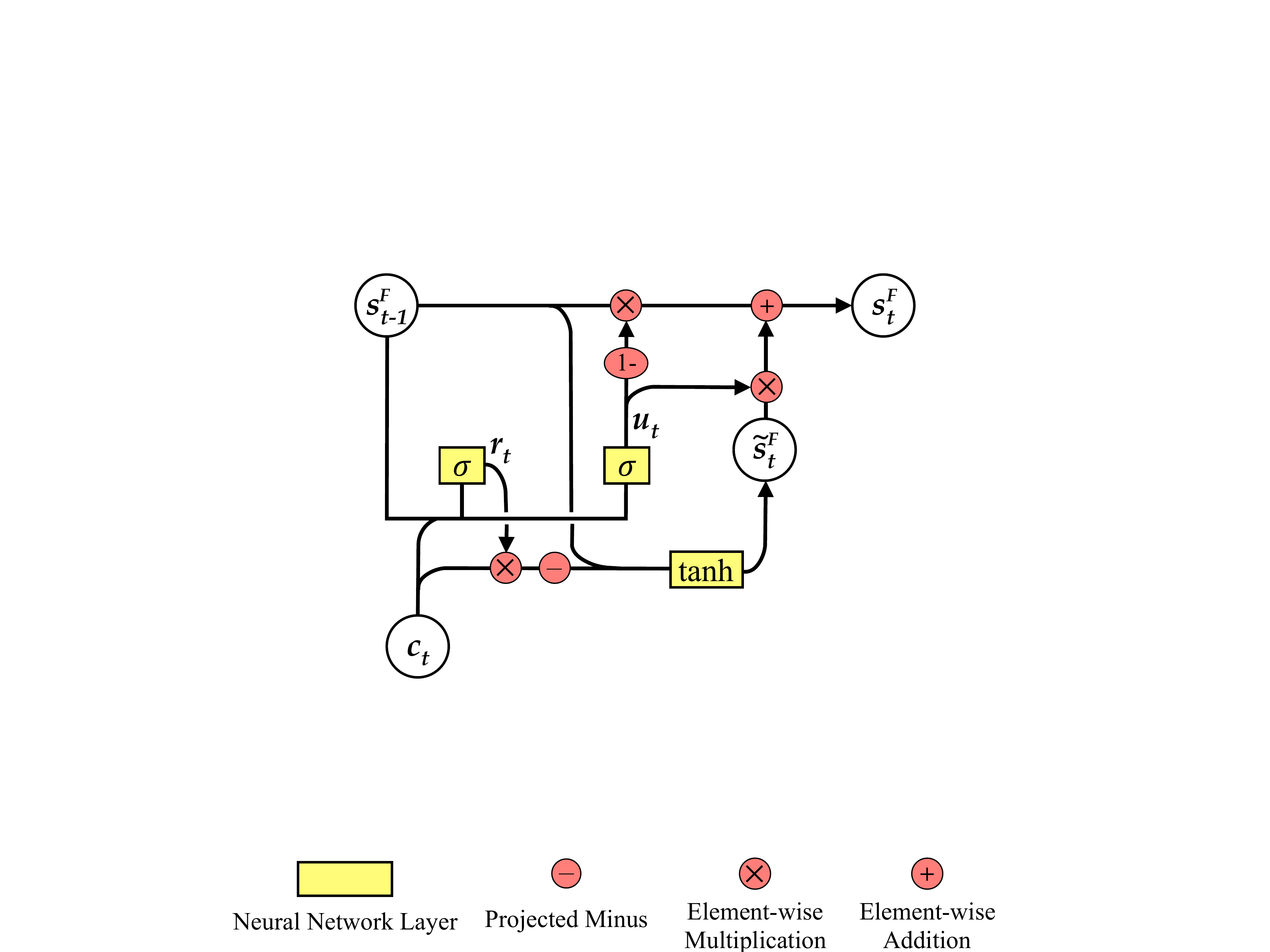}}
		\caption{Variants of activation functions for the \textsc{Future} layer.}
		\label{fig:activation}
	\end{figure*}

	\section{Modeling \textsc{Past} and \textsc{Future} for Neural Machine Translation}
	
	In this section, we describe how to separate \textsc{Past} and \textsc{Future} functions from decoding states. We introduce two additional RNN layers (Figure~\ref{fig:tri-sRNN}):
	\begin{compactitem}
		\item \textsc{Future} Layer (Section~\ref{sec-future}) encodes source contents to be translated.
		\item \textsc{Past} Layer (Section~\ref{sec-past}) encodes translated source contents.
	\end{compactitem}
	
	Let us take $\mathbf{y}=\{y_1, y_2, y_3, y_4\}$ as an example of the target sentence. The initial state of \textsc{Future} layer is a summarization of the whole source sentence, indicating that all source contents need to be translated.
	The initial state of \textsc{Past} layer is a all-zero vector, indicating no source content is yet translated.
	
	After $\mathbf{c}_1$ is obtained by the attention mechanism, we (1) update the \textsc{Future} layer by ``subtracting'' $\mathbf{c}_1$ from the previous state, and (2) update the \textsc{Past} layer state by ``adding'' $\mathbf{c}_1$ to the previous state.
	The two RNN states are updated as described above at every step of generating $y_1$, $y_2$, $y_3$, and $y_4$.
	In this way, at each time step, the \textsc{Future} layer encodes source contents to be translated in the future steps, while the \textsc{Past} layer encodes translated source contents up to the current step.
	
	The advantages of \textsc{Past} and \textsc{Future} layers are two-fold. First, they provide coverage information, which is fed to the attention model and guides NMT systems to pay more attention to untranslated source contents. Second, they provide a holistic view of the source information, since we would anticipate ``\textsc{Past} + \textsc{Future} = \textsc{Holistic}.''
	We describe them in detail in the rest of this section.

	\subsection{Modeling \textsc{Future}}
	\label{sec-future}
	Formally, the \textsc{Future} layer is a recurrent neural network~(the first gray layer in Figure \ref{fig:tri-sRNN})
	, and its state at time step $t$ is computed by
	\begin{equation}
	\mathbf{s}^{F}_t = \mathbb{F}(\mathbf{s}^{F}_{t-1}, \mathbf{c}_t)
	\label{eqn-future-state}
	\end{equation}
	where $\mathbb{F}$ is the activation function for \textsc{Future} layer. We have several variants $ \mathbb{F} $, aiming to better model the expected subtraction, as described in Section~\ref{sec-subtractive}.
	The \textsc{Future} RNN is initialized with the summarization of the whole source sentence, as computed by Equation~\ref{eqn-initialization}.
	
	When calculating attention context at time step $t$, we feed the attention model with the \textsc{Future} state from the last time step, which encodes source contents to be translated. We rewrite Equation~\ref{eqn-alignment-probability} as
	\begin{equation}
	\alpha_{t,i} = \mathrm{softmax}\big(a(\mathbf{s}_{t-1}, \mathbf{h}_i, \mathbf{s}^F_{t-1})\big)
	\end{equation}
	
	After obtaining attention context $\mathbf{c}_t$, we update \textsc{Future} states via Equation~\ref{eqn-future-state}, and feed both of them to decoder states:
	\begin{align}\label{eq:bi-sRNN}
	\mathbf{s}_t&=f(\mathbf{s}_{t-1}, y_{t-1}, \mathbf{c}_t, \mathbf{s}_t^F)
	\end{align}
	where $\mathbf{c}_t$ encodes the source context of the present translation, and $\mathbf{s}_t^F$ encodes source context on future translation.


	\subsubsection{Activation Functions for Subtraction}
	\label{sec-subtractive}
	
	We design several variants of RNN activation functions to better model the subtractive operation (Figure \ref{fig:activation}):
	
	\paragraph{GRU.}
	A natural choice is standard GRU,\footnote{Our work focuses on GRU, but can be applied to any RNN architectures such as LSTM.} which learns subtraction directly from the data:
	\begin{align}
	\mathbf{s}_t^F&=\mathrm{GRU}(\mathbf{s}_{t-1}^F, \mathbf{c}_t)\\
	&=\mathbf{u}_t \cdot \mathbf{s}_{t-1}^F + (1-\mathbf{u}_t) \cdot \mathbf{\tilde{s}}_t^F \nonumber\\
	\label{eq:gru_shat}
	\mathbf{\tilde{s}}_t^F &= \mathrm{tanh}(U(\mathbf{r}_t \cdot \mathbf{s}_{t-1}^F) + W\mathbf{c}_t)\\
	\label{eq:reset_gate}
	\mathbf{r}_t&= \sigma(U_r\mathbf{s}_{t-1}^F + W_r\mathbf{c}_t) \\
	\label{eq:update_gate}
	\mathbf{u}_t&= \sigma(U_u\mathbf{s}_{t-1}^F + W_u\mathbf{c}_t)
	\end{align}
	where $ \mathbf{r}_t $ is a reset gate determining the combination of the input with the previous state, and $ \mathbf{u}_t $ is an update gate defining how much of the previous state to keep around.
	The standard GRU uses a feed-forward neural network (Equation~\ref{eq:gru_shat}) to model the subtraction without any explicit operation, which may lead to the difficulty of the training.
	
	In the following two variants, we provide GRU with explicit subtraction operations, which are inspired by the well known phenomenon that minus operation can be applied to the semantics of word embeddings \cite{Mikolov2013}.\footnote{$\mathrm{E}(``\text{King}")-\mathrm{E}(``\text{Man}")=\mathrm{E}(``\text{Queen}")-\mathrm{E}(``\text{Woman}")$, where $\mathrm{E}(\cdot)$ is the embedding of a word.}
	Therefore we subtract the semantics being translated from the untranslated \textsc{Future} contents at each decoding step.

	\paragraph{GRU with Outside Minus (GRU-$\textbf{\textit{o}}$).}
	
	Instead of directly feeding $\mathbf{c}_t$ to GRU, we compute the current untranslated contents $\mathrm{M}(\mathbf{s}_{t-1}^F, \mathbf{c}_t)$ with an explicit minus operation, and then feed it to GRU:
	\begin{eqnarray}
	\mathbf{s}_{t}^F=\mathrm{GRU}(\mathbf{s}_{t-1}^F, \mathrm{M}(\mathbf{s}_{t-1}^F, \mathbf{c}_t))\\
	\mathrm{M}(\mathbf{s}_{t-1}^F, \mathbf{c}_t)=\tanh(   U_m\mathbf{s}_{t-1}^F-W_m\mathbf{c}_t)
	\end{eqnarray}

	\paragraph{GRU with Inside Minus (GRU-$\textbf{\textit{i}}$).}
	We can alternatively integrate a minus operation into the calculation of $\mathbf{\tilde{s}}_t^F$:
	\begin{equation}
	\tilde{\mathbf{s}}_t^F = \tanh(U\mathbf{s}_{t-1}^F-W(\mathbf{r_t} \cdot \mathbf{c}_t))
	\label{eqn-gru-i}
	\end{equation}
	Compared with Equation \ref{eq:gru_shat}, the differences between GRU-${\textit{i}}$ and standard GRU are
	\begin{compactenum}
		\item Minus operation is applied to produce the energy of intermediate candidate state $\tilde{\mathbf{s}}_t^F$;
		\item The reset gate $\mathbf{r}_{t}$ is used to control the amount of information flowing from inputs instead of from the previous state $\mathbf{s}_{t-1}^F$.
	\end{compactenum}
	
	\vspace{5pt}
	\noindent Note that for both GRU-$o$ and GRU-$i$, we leave enough freedom for GRU to decide the extent of integrating with subtraction operations. In other words, the information subtraction is ``soft.''

	\subsection{Modeling \textsc{Past}}
	\label{sec-past}
	Formally, the \textsc{Past} layer is another recurrent neural network (the second gray layer in Figure \ref{fig:tri-sRNN}), and its state at time step $t$ is calculated by
	\begin{equation}
	\mathbf{s}^{P}_t = \mathrm{GRU}(\mathbf{s}^{P}_{t-1}, \mathbf{c}_t)
	\label{eqn-past-state}
	\end{equation}
	Initially, $\mathbf{s}^{P}_t$ is an all-zero vector, which denotes no source content is yet translated. We choose $\mathrm{GRU}$ as the activation function for the \textsc{Past} layer, since the internal structure of $\mathrm{GRU}$ is in accord with ``addition'' operation.
	
	We feed the \textsc{Past} state from last time step to both attention model and decoder state:
	\begin{eqnarray}
	\alpha_{t,i} = \mathrm{softmax}\big(a(\mathbf{s}_{t-1}, \mathbf{h}_i, \mathbf{s}^P_{t-1})\big) \\
	\mathbf{s}_t = f(\mathbf{s}_{t-1}, y_{t-1}, \mathbf{c}_t,  \mathbf{s}^P_{t-1})
	\end{eqnarray}

	
\subsection{Modeling \textsc{Past} and \textsc{Future}}
     We integrate \textsc{Past} and \textsc{Future} layers together in our final model (Figure \ref{fig:tri-sRNN}):
    \begin{eqnarray}
	\alpha_{t,i} = \mathrm{softmax}\big(a(\mathbf{s}_{t-1}, \mathbf{h}_i, \mathbf{s}^F_{t-1}, \mathbf{s}^P_{t-1})\big) \\
	\mathbf{s}_t = f(\mathbf{s}_{t-1}, y_{t-1}, \mathbf{c}_t, \mathbf{s}_{t-1}^F, \mathbf{s}^P_{t-1})
	\end{eqnarray}
	In this way, both of the attention model and decoder state are aware of what has been translated, and what has not yet.
	
	\subsection{Learning}
	We introduce additional loss functions to estimate the semantic subtraction and addition, which guide the training of the \textsc{Future} layer and \textsc{Past} layer, respectively.
	\paragraph{Loss Function for Subtraction.}
	\label{sec-constraint}
	As described above, the \textsc{Future} layer models the future semantics in a declining way: $\mathbf{\Delta}_{t}^{F}=\mathbf{s}_{t-1}^F-\mathbf{s}_{t}^F \approx \mathbf{c}_t$. Since source and target sides contain equivalent semantic information in machine translation~\cite{Tu:2017:TACL}: $\mathbf{c}_t \approx \mathrm{E}(y_t)$, we directly measure the consistence between $\mathbf{\Delta}_{t}^{F}$ and $\mathrm{E}(y_t)$, which guides the subtraction to learn the right thing:
	\begin{eqnarray}\label{eq:cons}
	& loss(\mathbf{\Delta}_{t}^{F}, \mathrm{E}(y_t)) = -\log \frac{\exp\big(l(\mathbf{\Delta}_{t}^{F}, \mathrm{E}(y_t))\big)}{\sum_y \exp\big(l(\mathbf{\Delta}_{t}^{F}, \mathrm{E}(y))\big)} \nonumber \\
    & l(\mathbf{u}, \mathbf{v}) = \mathbf{u}^\top W \mathbf{v} + b\nonumber
	\end{eqnarray}
	In other words, we explicitly guide the \textsc{Future} layer by this subtractive loss, expecting $\mathbf{\Delta}_t^F$ to be discriminative of the current word $y_t$.

	\paragraph{Loss Function for Addition.}
	Likewise, we introduce another loss function to measure the information incrementation of the \textsc{Past} layer. Notice that  $\mathbf{\Delta}_{t}^{P}=\mathbf{s}_{t}^P-\mathbf{s}_{t-1}^P \approx \mathbf{c}_t $, which is defined similar to $\mathbf{\Delta}_t^F$ except a minus sign. In this way, we can reasonably assume the \textsc{Future} and \textsc{Past} layers are indeed doing subtraction and addition, respectively.
	
	\paragraph{Training Objective.}
	We train the proposed model $\Hat{\theta}$ on a set of training examples $\{\left[{\bf x}^n, {\bf y}^n\right]\}_{n=1}^{N}$, and the training objective is
	\begin{eqnarray}
	\Hat{\theta} = \argmin_{\theta}  \sum_{n=1}^{N} \sum_{t=1}^{|\mathbf{y}|} \bigg\{
	\underbrace{-\log P(y_t|y_{<t},\mathrm{\mathbf{x}}; \theta)}_\text{\normalsize \em neg. log-likelihood}  \nonumber \\
	+ \underbrace{loss(\mathbf{\Delta}_{t}^{F}, \mathrm{E}(y_t)|\theta)}_\text{\normalsize \em \textsc{Future} loss}  \nonumber \\
	+ \underbrace{loss(\mathbf{\Delta}_{t}^{P}, \mathrm{E}(y_t)|\theta)}_\text{\normalsize \em \textsc{Past} loss} \bigg\} \nonumber
	\end{eqnarray}
	

	\section{Experiments}
	
	\begin{table*}[ht]
		\centering
		\begin{tabular}{c|l|c|llll|lc}
			\toprule
			\#   & {\bf Model} & Dev & MT02 & MT04 & MT05 & MT06 & Avg.& $\Delta$\\
			\hline
			0   &   \textsc{RNNSearch}    &    35.90   &   36.84	  &   37.16   &   34.17   &   31.56   &   35.13   & -	\\
			\hline
			1   &   + \textsc{Frnn} (GRU)               &    36.11   &   36,94   &   38.52   &   34.58   &   32.08   &   35.65   &   +0.52  \\
			2   &   + \textsc{Frnn} (GRU-$o$)           &    36.70   &   37.81   &   38.59   &   35.10   &   32.60   &   36.16   &   +1.03  \\
			3   &   + \textsc{Frnn} (GRU-$i$)           &    36.98   &   38.24   &   38.66   &   34.68   &   32.66   &   36.24   &   +1.12  \\
			4   &   + \textsc{Frnn} (GRU-$i$) + \textsc{Loss}        &   37.15	 &   38.80   &	 39.13	 &	 35.79	 &   33.75	 &	 36.92	 &   +1.80 \\
			\hline
			5   &   + \textsc{Prnn}                     &	 36.90   &   37.62	 &	 39.04	 &	 35.24	 &	 32.80	 &	 36.32	 &   +1.19   \\
			6   &   + \textsc{Prnn} + \textsc{Loss}     &	 36.95	 &   39.06	 &	 39.55	 &	 35.05	 &	 33.80	 &   36.88	 &   +1.76   \\
			\hline
			7   &   \tabincell{l}{+ \textsc{Frnn} (GRU-$i$) + \textsc{Prnn}}     &	 37.44	 &   37.26   &	 39.10	 &	 35.29	 &	 32.78	 &	 36.37	&   +1.25\\
			8   &   \tabincell{l}{+ \textsc{Frnn} (GRU-$i$) + \textsc{Prnn} + \textsc{Loss}} & \textbf{37.90} & \textbf{39.65} & \textbf{40.37} & \textbf{36.75} & \textbf{34.55} & \textbf{37.84}  & \textbf{+2.71}\\
			\hline
			\hline
			9   & \textsc{RNNSearch-2dec}                        & 35.56 &   36.74   &  37.38    &   34.09   &   31.82   &   35.12   &   -0.01	\\
			10   & \textsc{RNNSearch-3dec}                       & 36.07 &   37.64	 &  37.62    &   34.14   &   32.73   &   35.64   &   +0.51	\\
			11   & \textsc{Coverage}~\cite{tu-EtAl:2016:P16-1}   & 36.56 &   37.54   &  38.39    &   34.47   &   32.38   &   35.87   &   +0.74 \\
			\bottomrule
		\end{tabular}
		\caption{Case-insensitive BLEU on Chinese-English Translation. ``\textsc{Loss}" means applying loss functions for \textsc{Future} layer (\textsc{Frnn}) and \textsc{Past} layer (\textsc{Prnn}). }
		\label{tab:Zh-En_bleu}
	\end{table*}

	\paragraph{Dataset.}
	We conduct experiments on  Chinese-English~(Zh-En), German-English~(De-En), and English-German~(En-De) translation tasks.
	
	For Zh-En, the training set consists of 1.6m sentence pairs, which are extracted from the LDC corpora\footnote{The corpora includes LDC2002E18, LDC2003E07, LDC2003E14, Hansards portion of LDC2004T07, LDC2004T08 and LDC2005T06}.
	The NIST 2003 (MT03) dataset is our development set; the NIST 2002 (MT02), 2004 (MT04), 2005 (MT05), 2006 (MT06) datasets are test sets.
	We also evaluate the alignment performance on the standard benchmark of \newcite{Liu2015Contrastive}, which contains 900 manually aligned sentence pairs. We measure the alignment quality with the alignment error rate~\cite{Och2003}.
	
	For De-En and En-De, we conduct experiments on the WMT17~\cite{WMT:2017} corpus. The dataset consists of 5.6M sentence pairs. We use \texttt{newstest2016} as our development set, and \texttt{newstest2017} as our testset.
	We follow \newcite{edinWMT17:arxiv} to segment both German and English words into subwords using byte-pair encoding~\cite[BPE]{Sennrich2016Neural}.
	
	We measure the translation quality with BLEU scores~\cite{papineni2002bleu}.
	We use the \texttt{multi-bleu} script for Zh-En~\footnote{\url{https://github.com/moses-smt/mosesdecoder/blob/master/scripts/generic/multi-bleu.perl}}, and the \texttt{multi-bleu-detok} script for De-En and En-De~\footnote{\url{https://github.com/EdinburghNLP/nematus/blob/master/data/multi-bleu-detok.perl}}.
	
	\paragraph{Training Details.}
	We use the  \texttt{Nematus}~\footnote{\url{https://github.com/EdinburghNLP/nematus}}~\cite{sennrich-EtAl:2017:EACLDemo}, implementing a baseline translation system, \textsc{RNNSearch}.
	For Zh-En, we limit the vocabulary size to 30K.
	For De-En and En-De, the number of joint BPE operations is 90,000.
	We use the total BPE vocabulary for each side.
	
	We tie the weights of the target-side embeddings and the output weight matrix~\cite{Press:EACL} for De-En.
	All out-of-vocabulary words are mapped to a special token \texttt{\small UNK}.
	
	We train each model with sentences of length up to 50 words in the training data. The dimension of word embeddings is 512, and all hidden sizes are 1024.
	In training, we set the batch size as 80 for Zh-En, and 64 for De-En and En-De.
	We set the beam size as 12 in testing.
	We shuffle the training corpus after each epoch.
	
	We use Adam~\cite{KingmaB14:adam:iclr} with annealing \cite{DBLP:journals/corr/DenkowskiN17} as our optimization algorithm. We set the initial learning rate as 0.0005, which halves when the validation cross-entropy does not decrease.
	
	
	%
	%
	
	For the proposed model, we use the same setting with the baseline model. 
	The \textsc{Future} and \textsc{Past} layer sizes are 1024.
	We employ a two-pass strategy for training the proposed model, which has proven useful to ease training difficulty when the model is relatively complicated~\cite{Shen:2016:ACL,Wang:2017:AAAI,Wang:2018:AAAI}.
	Model parameters shared with the baseline are initialized by the baseline model.
	

	\subsection{Results on Chinese-English}
	
	We first evaluate the proposed model on the Chinese-English translation and alignment tasks.
	
	\subsubsection{Translation Quality}
	
	Table~\ref{tab:Zh-En_bleu} shows the translation performances on Chinese-English.
	Clearly the proposed approach significantly improves the translation quality in all cases, although there are still considerable differences among different variants.
	
	\paragraph{\textsc{Future} Layer.} (Rows 1-4).
	All the activation functions for the \textsc{Future} layer obtain BLEU socre improvements: GRU +0.52, GRU-$o$ +1.03, and GRU-$i$ +1.12.
	Specifically, GRU-$o$ is better than a regular GRU for its minus operation, and GRU-$i$ is the best, which shows that our elaborately designed architecture is more proper for modeling the decreasing phenomenon of  the future semantics.
	
	Adding subtractive loss gives an extra 0.68 BLEU score improvement, which indicates that adding g is beneficial guided objective for \textsc{Frnn} to learn the minus operation.
	
	
	\paragraph{\textsc{Past} Layer.} (Rows 5-6). We observe the same trend on introducing \textsc{Past} layer: using it alone achieves a significant improvement~( +1.19), and with the additional objective it further improves the translation performance~( +0.57).

	\paragraph{Stacking \textsc{Future} and \textsc{Past} Together.} (Rows 7-8).
	The model's final  architecture outperforms our intermediate models (1-6) by combining \textsc{Frnn} and \textsc{Prnn}.
	By further separating the functionaries of past contents modeling and language modeling into different neural components, the final model is more flexible, obtaining a 0.91 BLEU  improvement over the best intermediate model (Row 4) and an improvement of 2.71 BLEU points over the \textsc{RNNSearch} baseline.
	
	\paragraph{Comparison with Other Work.} (Rows 9-11).
	We also conduct experiments with multi-layer decoders~\cite{wu2016google} to see whether NMT system can automatically model the translated and untranslated contents with additional decoder layers  (Rows 9-10).
	However, we find that the performance is not improved using a two-layer decoder (Row 9), until a deeper version (three-layer decoder, Row 10) is used.
	This indicates that enhancing performance is non-trivial by simply adding more RNN layers into the decoder without any explicit instruction, which is consistent with the observation of \newcite{DBLP:BritzGLL17}
	
	Our model also outperforms the word-level \textsc{Coverage}~\cite{tu-EtAl:2016:P16-1}, which considers the coverage information of the source words independently.
	Our proposed model can be regarded as a high-level coverage model, which captures higher level coverage information, and gives more specific signals for the decision of attention and target prediction.
	Our model is more deeply involved in generating target words, by being fed not only to the attention model as in~\newcite{tu-EtAl:2016:P16-1}, but also to the decoder state.
	

	\begin{table*}[t]
		\setcounter{table}{4}
		\centering
		\begin{tabular}{c|l|cc|cc}
		\toprule
			\multirow{2}{*}{\bf System} &	 \multirow{2}{*}{\bf Architecture}		&	\multicolumn{2}{c|}{\bf De-En}          &   \multicolumn{2}{c}{\bf En-De}\\
			\cline{3-6}
			&  &  Dev &  Test         &   Dev    &   Test\\
			\hline
			\multirow{2}{*}{\newcite{Del2017:wmt17}}   & cGRU + BPE + dropout                                & 31.9      & 27.2                      & 27.4      & 21.0     \\
			
			& ~~~~~~+ name entity forcing + {\em synthetic data}     & 36.9      & 29.0                      & 30.9      & 22.7     \\
			\hline
			\multirow{2}{*}{\newcite{escolan:2017:WMT}}              & Char2Char + Rescoring with inverse model       & 32.1      & -                         & 27.0      & -     \\
			& ~~~~~~+ {\em synthetic data}     & -      & 28.1                      & -      &  21.2   \\
			\hline
			\newcite{edinWMT17:arxiv}               & cGRU + BPE + {\em synthetic data}                    & 38.0      & 32.0                      & 32.2      & 26.1     \\
			\hline
			\multirow{3}{*}{\em this work}  &   \textsc{Base}                       & 32.0      & 27.8                      & 28.3      & 23.3     \\
			&   \textsc{Coverage}                                               & 32.2      & 28.7                      & 28.9      & 23.6     \\
			&   \textsc{Ours}                                                   & 33.5      & 29.7                      & 29.5      & 24.3 \\
		\bottomrule
		\end{tabular}
		\caption{Results of  De-En and En-De ``{\em synthetic data}'' denotes additional 10M monolingual sentences, which is not used in our work.}
		\label{tab:De-En_bleu}
	\end{table*}

		\begin{table}[t]
		\setcounter{table}{2}
		\centering
		\begin{tabular}{l|c|c|c|c}
		\toprule
			\multirow{2}{*}{\bf Model}			&	\multicolumn{2}{c|}{\bf Over-Trans}   &   \multicolumn{2}{c}{\bf Under-Trans}\\
			\cline{2-5}
			&   Ratio    &   $\Delta$    &   Ratio    &   $\Delta$\\
			\hline
			\textsc{Base}                       &   1.7\%   &   --         &   8.8\%   &   --      \\
			\textsc{Coverage}                   &   1.5\%   &   -11.8\%     &   7.7\%   &   -12.4\% \\
			\textsc{Ours}                       &   1.6\%   &   -5.9\%    &   5.7\%   &   -35.2\% \\
		\bottomrule
		\end{tabular}
		\caption{Subjective evaluation on over- and under-translation for Chinese-English. ``Ratio'' denotes the percentage of source words which are over- or under-translated, ``$\Delta$'' indicates relative improvement. ``\textsc{Base}'' denotes \textsc{RNNSearch} and ``\textsc{Ours}'' denotes ``+ \textsc{Frnn} (GRU-$i$) + \textsc{Prnn} + \textsc{Loss}''.}
		\label{table-subjective-evaluation}
	\end{table}

	\subsubsection{Subjective Evaluation}
	Following~\newcite{tu-EtAl:2016:P16-1}, we conduct subjective evaluations to validate the benefit of modeling \textsc{Past} and \textsc{Future}~(Table \ref{table-subjective-evaluation}). Four human evaluators are asked to evaluate the translations of 100 source sentences, which are randomly sampled from the testsets without knowing from which system the translation is selected.
	For the \textsc{Base} system, 1.7\% of the source words are over-translated and 8.8\% are under-translated.
	Our proposed model alleviates these problems by explicitly modeling the dynamic source contents by \textsc{Past} and \textsc{Future} layers, reducing 11.8\% and 35.2\% of over-translation and under-translation errors, respectively. The proposed model is especially effective for alleviating the under-translation problem, which is a more serious translation problem for NMT systems, and is mainly caused by lacking necessary coverage information~\cite{tu-EtAl:2016:P16-1}.

	\subsubsection{Alignment Quality}


	Table~\ref{table-alignment-results} lists the alignment performances of our proposed model.
	We find that the \textsc{Coverage} model do improve attention model.
   But our model can produce much better alignments compared to the word level coverage~\cite{tu-EtAl:2016:P16-1}.
	Our model distinguishes the \textsc{Past} and \textsc{Future} directly, which is a higher level coverage mechanism than the word coverage model.

	\begin{table}[t]
		\setcounter{table}{3}
		\centering
		\begin{tabular}{l|c|c}
		\toprule
			{\bf Model}			    &	{\bf AER}   &   $\Delta$\\
			\hline
			\textsc{Base}           &   39.73       &   --\\
			\textsc{Coverage}       &   38.73       &   -1.00\\
			\textsc{Ours}           &   35.90       &   -3.83\\
		\bottomrule
		\end{tabular}
		\caption{Evaluation of the alignment quality. The lower the score, the better the alignment quality.}
		\label{table-alignment-results}
	\end{table}

	\subsection{Results on German-English}

	
	We also evaluate our model on the WMT17 benchmarks for both De-En and En-De.
	As shown in Table \ref{tab:De-En_bleu}, our baseline gives comparable BLEU scores to the state-of-the-art NMT systems of WMT17.
	Our proposed model improves the strong baseline on both De-En and En-De.
	This shows that our proposed model work well across different language pairs.
	\newcite{Del2017:wmt17} and \newcite{edinWMT17:arxiv} obtain higher BLEU scores than our model, because they use additional large scaled synthetic data~(about 10M) for training.
	It maybe unfair to compare our model to theirs directly.

	\subsection{Analysis}
	
	We conduct analyses on Zh-En, to better understand our model from different perspectives .
	
	\paragraph{Parameters and Speeds.}
	As shown in Table \ref{tab:speed}, the baseline model (\textsc{Base}) has 80M parameters. A single \textsc{Future} or \textsc{Past} layer introduces 15M to 17M parameters, and the corresponding objective introduces 18M parameters. In this work, the most complex model introduces 65M parameters, which leads to a relatively slower training speed.
	However, our proposed model does not significant slow down the decoding speed. The most time consuming part is the calculation of the subtraction and addition losses. As we show in the next paragraph, our system works well by only using the losses in training, which further improve the decoding speed of our model.

	\begin{table}[t]
		\setcounter{table}{5}
		\centering
		\begin{tabular}{l|r|cc}
			\toprule
			\multirow{2}{*}{\bf Model}	&	\multirow{2}{*}{\bf \#Para.} &	\multicolumn{2}{c}{\bf Speed}\\
			\cline{3-4}                 &                                &  Train   &   Test\\
			\hline
			\textsc{Base}                             &   80M      &  42.59    &   2.05\\
			\hline
			+ \textsc{Frnn} (GRU)                           &   96M      &  31.91    &   1.99\\
			+ \textsc{Frnn} (GRU-$o$)                       &   97M      &  30.88    &   1.93\\
			+ \textsc{Frnn} (GRU-$i$)                       &   97M      &  31.06    &   1.95\\
			~~~~~ + \textsc{Loss}                           &   115M     &  29.40    &   1.68\\
			\hline
			+ \textsc{Prnn}                                 &   95M      &  32.01    &   1.98\\
			~~~~~ + \textsc{Loss}                           &   113M     &  29.60    &   1.69\\
			\hline
			+ \textsc{Frnn} + \textsc{Prnn}                 &   110M     &  26.01    &   1.88\\
			~~~~~ + \textsc{Loss}                           &   145M     &  22.94    &   1.52\\
			\hline
			\hline
			\textsc{RNNSearch-2dec}                         &   93M      &  39.89    &   2.00\\
			\textsc{RNNSearch-3dec}                         &   105M     &  35.57    &   1.82\\
			\textsc{Coverage}                               &   80M      &  40.48    &   1.90\\
			\bottomrule
		\end{tabular}
		\caption{Statistics of parameters, training and testing speeds (sentences per second).}
		\label{tab:speed}
	\end{table}

	

	\paragraph{Effectiveness of Subtraction and Addition Loss.}

	Adding subtraction and addition loss functions helps in twofold: (1) guiding the training of the proposed subtraction and addition operation, and (2) enabling better reranking of generated candidates in testing. Table~\ref{table-contribution} lists the improvements from the two perspectives.
	When applied only in training, the two loss functions lead to an improvement of 0.48 BLEU points by better modeling subtraction and addition operations.
	On top of that, reranking with \textsc{Future} and \textsc{Past} loss scores in testing further improves the performance by +0.99 BLEU points.

	
	\begin{table}[t]
		\centering
		\begin{tabular}{c|cc|cc}
		\toprule
			\multirow{2}{*}{\bf Model}  &   \multicolumn{2}{c|}{\bf \textsc{Loss} used in}	&	\multirow{2}{*}{\bf BLEU}  & \multirow{2}{*}{\bf $\Delta$}\\
			\cline{2-3}
			&   {\em Train}	&	{\em Test}	&	\\
			\hline
			\textsc{Base}   &   --  &   --  &   35.13   &  --\\
			\hline
			\multirow{3}{*}{\textsc{Ours}}  &   \texttimes	&	\texttimes	&	36.37	&   +1.25\\
			&   \checkmark	&	\texttimes	&	36.85   &   +1.72\\
			&   \checkmark	&	\checkmark	&	37.84   &   +2.71\\
		\bottomrule
		\end{tabular}
		\caption{Contributions of loss functions from parameter training (``{\em Train}'') and reranking of candidates in testing (``{\em Test}'').}
		\label{table-contribution}
	\end{table}

	\paragraph{Initialization of \textsc{Future} Layer.}
	
	\begin{table}[t]
		\centering
		\begin{tabular}{l|c}
		\toprule
			{\bf Initialize \textsc{Frnn} with \dots} & {\bf BLEU}\\
			\hline
			Source Summarization    & 36.24\\
			All-Zero Vector         & 35.81\\
		\bottomrule
		\end{tabular}
		\caption{Influence of initialization of \textsc{Frnn} layer~(GRU-$i$) }
		\label{table-future-init}
	\end{table}
	
	The baseline model does not obtain abundant accuracy improvement by feeding the source summarization into the decoder~(Table \ref{tab:bleu_init}).  We also experiment to not feed the source summarization into the decoder of the proposed model, which leads to a significant BLEU score drop on Zh-En.
	This shows that our proposed model better use the source summarization with explicitly modeling the \textsc{Future} compared to the conventional encoder-decoder baseline.

	%
	%
	%
	\begin{CJK}{UTF8}{gbsn}
		\begin{table*}[h]
			\renewcommand\arraystretch{1.2}
			\centering
			\begin{tabular}{m{57pt}|m{380pt}}
				\toprule
				Source & 布什 还 表示 , 应 巴基斯坦 和 印度 政府 的 邀请 , 他 将 于 3月份 对 巴基斯坦 和 印度 进行 访问 。\\
				\hline
				Reference & bush also said that {\bf \color{red} at the invitation of the pakistani and indian governments }, he would visit pakistan and india in march .\\
				\hline
				\textsc{Base} & bush also said that he would visit pakistan and india in march .\\
				\hline
				\textsc{Coverage} & bush also said that at the invitation of pakistan and india , he will visit pakistan and india in march .\\
				\hline
				\textsc{Ours} &  bush also said that at the invitation of the pakistani and indian governments , he will visit pakistan and india in march .\\
				\hline
				\hline
				Source & 所以 有 不少 人 认为 说 , 如果 是 这样 的 话 , 对 皇室 、 对 日本 的 社会 也 是 会 有 很 大 的 影响 的 。\\
				\hline
				Reference &  therefore , many people say that it will have a great impact on {\bf \color{red}the royal family} and japanese society .\\
				\hline
				\textsc{Base} &  therefore , many people are of the view that if this is the case , it will also have a great impact on {\em \color{blue} the people of hong kong} and the japanese society .\\
				\hline
				\textsc{Coverage} &  therefore , many people think that if this is the case , there will be great impact on the royal and japanese society .\\
				\hline
				\textsc{Ours} &   therefore , many people think that if this is the case , it will have a great impact on the royal and japanese society .\\
				\bottomrule
			\end{tabular}
			\caption{Comparison on Translation Examples. We italicize some {\em \color{blue} translation errors} and highlight the {\bf \color{red} correct ones} in bold.}
			\label{tab:case}
		\end{table*}
	\end{CJK}
	
	\paragraph{Case Study.}
	We also compare the translation cases for the baseline, word level coverage and our proposed models.
	As shown in Table \ref{tab:case}, our baseline system suffers from the over-translation problems (case 1), which is consistent with the results of human evaluation~(Section \ref{table-subjective-evaluation}). 
	The \textsc{Base} system also incorrectly translates ``the royal family'' into ``the people of hong kong'', which is totally irrelevant here. We attribute the former case to the lack of untranslated future modeling, and the latter one to the overloaded use of the decoder state where the language modeling of the decoder leads to the fluent but wrong predictions.
	In contrast, the proposed approach almost address the errors in these cases.

	\section{Conclusion}
	Modeling source contents well is crucial for encoder-decoder based NMT systems.
	However, current NMT models suffer from distinguishing translated and untranslated translation contents, due to the lack of explicitly modeling past and future translations.
	In this paper, we separate \textsc{Past} and \textsc{Future} functionalities from decoder states, which can maintain a dynamical yet holistic view of the source content at each decoding step.
	Experimental results show that the proposed approach significantly improves translation performances across different language pairs.
	With better modeling of past and future translations, our approach performs much better than the standard attention-based NMT, reducing the errors of under and over translations.

	\section{Acknowledgement}
	We would like to thank the anonymous reviewers for their insightful comments. Shujian Huang is the corresponding author. This work is supported by the National Science Foundation of China (No. 61672277, 61772261), the Jiangsu Provincial Research Foundation for Basic Research (No. BK20170074).

	\balance
	\bibliography{tacl}
	\bibliographystyle{acl2012}
	
\end{document}